\documentclass[11pt]{article}

\usepackage[preprint]{acl}

\usepackage{times}
\usepackage{latexsym}

\usepackage[T1]{fontenc}

\usepackage[utf8]{inputenc}

\usepackage{microtype}

\usepackage{inconsolata}

\usepackage{graphicx}

\usepackage{microtype}
\usepackage{booktabs}

\usepackage{inconsolata}
\usepackage{amsmath}
\usepackage{graphicx}
\usepackage{xcolor}
\usepackage{graphicx}
\usepackage{subcaption}
\usepackage{multirow}
\usepackage{booktabs}
\usepackage{subcaption}
\usepackage[font=small,labelfont=bf]{caption}
\usepackage{pifont} 

\usepackage[most]{tcolorbox}

\usepackage[most]{tcolorbox}
\usepackage{ragged2e}     
\usepackage{enumitem} 

\tcbset{
  onecolqual/.style={
    enhanced,
    breakable,
    width=\columnwidth,   
    colback=orange!4,
    colframe=black!25,
    boxrule=0.35pt,
    arc=2pt, outer arc=2pt,
    left=6pt,right=6pt,top=5pt,bottom=5pt,
    boxsep=0pt,
    before skip=6pt, after skip=6pt,
  }
}
\title{MathRobust-LV: Evaluation of Large Language Models' Robustness to Linguistic Variations in Mathematical Reasoning}

\author{Neeraja Kirtane\thanks{Equal contribution.}, \hspace{2pt} 
Yuvraj Khanna\footnotemark[1], \hspace{2pt}
Peter Relan \\ 
 Got It Education \\
{\texttt{\{neeraja,yuvraj,peter\}@gotiteducation.com}}
}

\begin{document}
\maketitle
\begin{abstract}
Large language models excel on math benchmarks, but their math reasoning robustness to linguistic variation is underexplored. While recent work
increasingly treats high-difficulty competitions like
the IMO as the gold standard for evaluating reasoning, we believe in comprehensive benchmarking of high school level math problems in real educational settings.
We introduce MathRobust-LV, a test set and evaluation methodology that mirrors how instructors rephrase problems across assessments while keeping difficulty constant: we change surface details (names, contexts, variables) while preserving numerical structure and answers. In contrast to prior efforts that alter problem content or emphasize IMO-level tasks, we focus on high-school-level dataset problems at the difficulty level where models are currently deployed in educational settings: tutoring and assessment systems. In these applications, instructors rephrase identical concepts in varied ways, making linguistic robustness essential for reliable deployment.
Although MATH data benchmarking is often regarded as saturated, our experiment on 34 models reveals that accuracy declines when moving from the baseline to the variants. 
These drops are severe for smaller models (9-11\%) while stronger models also show measurable degradation. 
Frontier models like \texttt{GPT-5}, \texttt{Gemini-2.5pro} remain comparatively stable.  Our results highlight that robustness to linguistic variation is a fundamental challenge, exposing reasoning vulnerabilities in models.
\end{abstract}

\section{Introduction}

Large Language Models (LLMs) have demonstrated impressive performance on a wide range of mathematical benchmarks, including GSM8K \cite{cobbe2021training} and MATH \cite{hendrycks2021measuring} and recently even on Olympiad-level problems \footnote{\url{https://huggingface.co/datasets/Maxwell-Jia/AIME_2024}}. These advances have led to the perception that top-performing models possess near-human mathematical reasoning abilities. While the models have improved in reasoning significantly, there remains a question as to how robust these models are. In this context, by robustness we refer to mathematical robustness of models. This implies that if a model is robust it will not falter when changes are made to the question and also maintain high baseline accuracy.
Emerging evidence suggests that high performance may not stem from robust reasoning but rather from superficial pattern matching, dataset leakage, or overfitting to training distributions \citep{zhou2022teaching, zhang2024careful}. 

Recent work by Apple introduced GSM-Symbolic \cite{mirzadeh2024gsm}, a benchmark that perturbs math problems by modifying surface features such as names, numbers, and intermediate steps, showing that LLMs are highly sensitive to symbolic or structural shifts. However, GSM-Symbolic simultaneously alters multiple aspects of the original problems, both content and solution making it difficult to isolate the source of performance degradation.
\begin{figure*}[ht]
    \centering
    \includegraphics[width=0.7\linewidth]{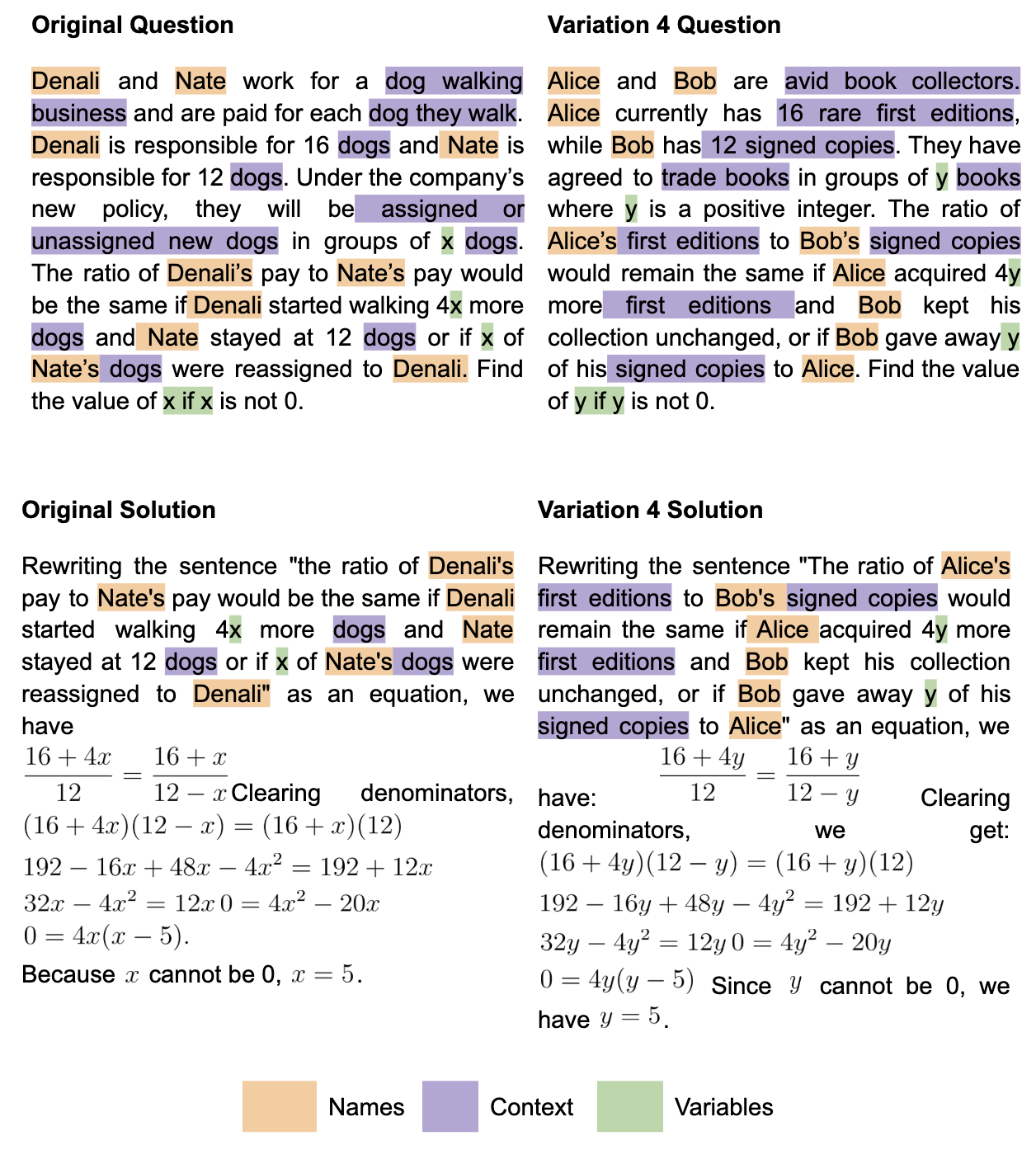}
    \caption{All the different types of substitutions done in a question to generate variations. Change are highlighted in different colors to show the different components that are changed.}
    \label{fig:variation4example}
\end{figure*}
In this paper, we present \textbf{MathRobust-LV (Linguistic Variation)}, a new methodology that probes LLM robustness through \emph{minimal linguistic perturbations}, while strictly preserving the problem’s numerical structure, symbolic logic, and correct final answer. We design our variations to mimic how instructors would vary their problems. They would keep the reasoning backbone the same and change the phrasing to ensure that the questions are different for all students, yet have the same level of difficulty. Therefore, the variations test the model’s ability to adapt to different phrasings of the same underlying mathematical problem, simulating real-world variation in how users might phrase identical tasks. A student who truly understands quadratic equations should solve them regardless of whether the problem asks about `load distribution' or `bridge arc shapes' Similarly, a model with robust mathematical reasoning should not fail when `variable x' becomes `variable y' or `Alice' becomes `Bob.' Our evaluation tests this basic consistency.


While recent work increasingly treats high-level competitions like the IMO as the gold standard for evaluating mathematical reasoning in LLMs, we deliberately focus on high-school-level problems because this difficulty level represents where models are currently deployed in real educational settings: tutoring systems, homework help platforms, and automated assessment tools. In these settings, students naturally rephrase problems in varied ways, and models must handle identical mathematical concepts expressed through different contexts and phrasings. If models cannot maintain consistency under such trivial rewordings, they lack the robustness required for reliable educational deployment, regardless of performance on competition benchmarks.

We initially focus our perturbations on four controlled axes: (1) variable substitution, (2) context substitution, (3) complete rephrasing with variables fixed, and (4) complete rephrasing with variable substitution. After running preliminary experiments to understand which variation is better to scale for our evaluations, we go ahead with Variation 4. We also choose variation 4 as it is a combination of all the previous three variations. More details are in 
(\S\ref{sec:experimental_design}).


We expect zero drops in the accuracy because the reasoning backbone is preserved. However, we observe drops of 2-11\%, suggesting models rely on surface features. When the baseline accuracy is low, there might be an increase in the variation accuracy, leading to a positive increase. But when we are testing for robustness, we want the base and variant accuracy to be high and also have no drop. 

We test the variations on 34 popular and widely used open and closed-source models, ranging from 0.6B to 70B parameters and even bigger closed source models. Our main contributions are as follows:
\begin{table*}[hbt!]
\centering

\begin{tabular}{lcccc}
\toprule
\textbf{Model} & \textbf{Params (B)} & \textbf{Baseline(130)} & \textbf{Variant(520)} & \textbf{Drop} \\
\cmidrule(lr){3-5}
 & & (\%) & (\%) & (\%) \\
\midrule
Qwen3-0.6B & 0.75 & 34.53 & 37.69 & +3.16 \\
\cmidrule(lr){1-5}
OpenMath-Nemotron-1.5B & 1.54 & 75.36 & 63.84 & \textcolor{red!60}{-11.52} \\
DeepScaleR-1.5B-Preview & 1.78 & 59.13 & 55.77 & \textcolor{red!60}{-3.36} \\
DeepSeek-R1-Distill-Qwen-1.5B & 1.78 & 58.44 & 48.85 & \textcolor{red!60}{-9.59} \\
\cmidrule(lr){1-5}
Qwen3-1.7B & 2.03 & 59.16 & 56.93 & \textcolor{red!60}{-2.23} \\
Phi-4-mini-instruct & 3.84 & 26.75 & 28.65 & +1.90 \\
Phi-4-mini-reasoning & 3.84 & 69.97 & 63.85 & \textcolor{red!60}{-6.13} \\
\cmidrule(lr){1-5}
Qwen3-4B & 4.02 & 82.28 & 77.12 & \textcolor{red!60}{-5.17} \\
Qwen3-4B-Thinking-2507 & 4.02 & 75.26 & 72.50 & \textcolor{red!60}{-2.77} \\
Llama-3.1-Nemotron-Nano-4B-v1.1 & 4.51 & 70.65 & 66.73 & \textcolor{red!60}{-3.92} \\
AceMath-RL-Nemotron-7B & 7.62 & 79.20 & 77.11 & \textcolor{red!60}{-2.09} \\
DeepSeek-R1-Distill-Qwen-7B & 7.62 & 73.08 & 67.12 & \textcolor{red!60}{-5.96} \\
OpenMath-Nemotron-7B & 7.62 & 83.05 & 82.12 & \textcolor{red!60}{-0.94} \\
\cmidrule(lr){1-5}
DeepSeek-R1-Distill-Llama-8B & 8.03 & 65.31 & 65.00 & \textcolor{red!60}{-0.31} \\
Llama-3.1-8B-Instruct & 8.03 & 21.39 & 24.23 & +2.83 \\
Llama-3.1-Nemotron-Nano-8B-v1 & 8.03 & 83.03 & 76.93 & \textcolor{red!60}{-6.10} \\
DeepSeek-R1-0528-Qwen3-8B & 8.19 & 68.39 & 67.50 & \textcolor{red!60}{-0.89} \\
Qwen3-8B & 8.2 & 77.62 & 79.42 & +1.80 \\
Phi-4 & 14.7 & 48.33 & 48.63 & +0.30 \\
Phi-4-reasoning & 14.7 & 81.54 & 78.46 & \textcolor{red!60}{-3.08} \\
Phi-4-reasoning-plus & 14.7 & 83.08 & 80.77 & \textcolor{red!60}{-2.31} \\
Qwen3-14B & 14.8 & 86.13 & 84.42 & \textcolor{red!60}{-1.71} \\
\cmidrule(lr){1-5}
Qwen3-30B-A3B & 30.5 & 84.56 & 83.46 & \textcolor{red!60}{-1.10} \\
Qwen3-30B-A3B-Thinking-2507 & 30.5 & 86.10 & 80.96 & \textcolor{red!60}{-5.14} \\
QwQ-32B & 32.5 & 84.62 & 82.50 & \textcolor{red!60}{-2.12} \\
QwQ-32B-Preview & 32.5 & 66.08 & 65.39 & \textcolor{red!60}{-0.69} \\
DeepSeek-R1-Distill-Qwen-32B & 32.8 & 79.98 & 78.85 & \textcolor{red!60}{-1.13} \\
Qwen3-32B & 32.8 & 89.21 & 84.81 & \textcolor{red!60}{-4.40} \\
\cmidrule(lr){1-5}
DeepSeek-R1-Distill-Llama-70B & 70.6 & 83.05 & 81.54 & \textcolor{red!60}{-1.52} \\
\midrule
\multicolumn{5}{l}{\textbf{Closed-source models}} \\
\cmidrule(lr){1-5}
Claude-sonnet & -- & 73.85 & 70.58 & \textcolor{red!60}{-3.27} \\
Claude-opus & -- & 75.38 & 72.11 & \textcolor{red!60}{-3.27} \\
Gemini-2.5pro & -- & 96.13 & 96.35 & +0.21 \\
Gemini-2.5flash & -- & 94.62 & 91.16 & \textcolor{red!60}{-3.46} \\
GPT-5 & -- & 97.69 & 96.92 & \textcolor{red!60}{-0.77} \\
\bottomrule
\end{tabular}
\caption{Baseline vs Variant results. The drop is calculated as the difference between the variant and baseline. A negative drop indicates that the model is not robust to variations. (Exact param size is taken from huggingface)}
\label{tab:final_table}
\end{table*}

\begin{itemize}
    \item We propose \textbf{MathRobust-LV}, a new methodology for evaluating linguistic robustness in math reasoning, grounded in minimal but meaning-preserving rewordings. We have a total of 520 variations for 130 original questions.
    \item We perform a large-scale evaluation of these variations across models of various types and sizes and report the performance drops in the accuracy.
    \item We discuss differences in accuracy across sizes and different families of models. We also discuss the difference in performance across two different data sources.
\end{itemize}



\section{Related Work}


There is a considerable amount of work being done to show that reasoning in LLMs is fragile. Benchmarks such as GSM-Plus \cite{li2024gsm} and MATH-Perturb \cite{huang2025math} extend GSM8K and MATH with controlled surface-form variations, showing that even minor modifications can cause substantial accuracy drops across a wide range of models. Other work has probed robustness to non-semantic noise, such as punctuation insertions in ArithmAttack \cite{abedin2025arithmattack}, or numerical perturbations in operands and constants \cite{yang2025evaluating}, revealing weaknesses in both arithmetic and reasoning steps. Frameworks like Math-RoB \cite{yu2025benchmarking} and AR-Checker \cite{hou2025automatic} perform stress testing via automatically generated semantically equivalent variants, while causal analysis approaches \cite{stolfo2022causal} quantify sensitivity to surface form, operator, and operand changes. 

More recently, Apple introduced GSM-Symbolic \cite{mirzadeh2024gsm}, which modifies both surface and symbolic structure including numbers and names highlighting sensitivity of LLMs to structural transformations. However, the method conflates linguistic and mathematical variation.
\citet{zhang2024careful} re-examined GSM8K by introducing GSM1K, exposing a performance drop from distribution shift. Similarly, MathOdyssey \citep{fang2024mathodyssey} and Big‑Math \citep{albalak2025big} extend robustness testing to higher difficulty domains, utilizing broader question distributions. \cite{hao2025investigation} investigate linguistic and parametric changes on Olympiad-level datasets. The work by \cite{yu2025benchmarking} also measures robustness of models on reasoning by making changes to the chains in the solution. The CORE framework by \cite{hong2024evaluating} shows a systematic way to dynamically alter various parts of a problem to generate perturbations. While they include linguistic perturbation as one of the changes, we study that in more detail and depth across more model sizes and types.

\paragraph{How our work differs.}
Unlike prior approaches: We preserve \emph{numerical values}, \emph{symbolic variables}, and \emph{final answers} exactly, isolating purely linguistic perturbations. We design variations that decouple surface phrasing from formal reasoning structure. We expect a zero drop in accuracy as the original reasoning structure is similar. We offer a large-scale evaluation across different models and find consistent accuracy degradation. Our evaluation method complements existing robustness suites, offering a targeted lens into the linguistic brittleness of LLMs in mathematical reasoning, when subjected to likely conditions in real education settings.
\begin{table}[ht]
\centering
\small
\begin{tabular}{lcc}
\toprule
\textbf{Subject} & \textbf{AoPS (\%)} & \textbf{MATH (\%)} \\
\midrule
Algebra                   & 1.6  & 49.2 \\
Counting \& Probability   & 56.5 & 9.2 \\
Geometry                  & 19.4 & 12.3 \\
Intermediate Algebra      & 4.8  & 9.2 \\
Number Theory             & 16.1 & 12.3 \\
Precalculus               & 1.6  & 7.7 \\
\bottomrule
\end{tabular}
\caption{Distribution of Level 4 \& 5 problems per subject in AoPS vs. MATH datasets (percentages).}
\label{tab:distributiontopic}
\end{table}

\section{Methodology Design}
\subsection{Seed Dataset Selection}
\label{sec:seed_dataset}
We utilize the \textbf{MATH}\cite{hendrycks2021measuring} dataset as one of the seed datasets to generate variations. The MATH dataset has more challenging problems as compared to GSM8k. It contains questions from various topics like prealgebra, algebra, counting and probability, geometry, intermediate algebra, precalculus, and number theory. This allows us to assess the given model better across various topics and difficulty levels. From this, we focus on problems of difficulty of Level 4 and 5. (Math problems in widely used datasets are calibrated on a difficulty level of 1-10.). To ensure variable generation, each problem should have a both a mathematical variable and a non-mathematical context. We this constraint, we select 65 problems from the MATH dataset.

In addition to the MATH data, we scrape problems of difficulty level 4 and 5 from past competition math questions\footnote{\url{https://artofproblemsolving.com/wiki/index.php/AoPS_Wiki:Competition_ratings}} that were not included in MATH train or test set. We ensure that the scraped problems have a mathematical context and a non-mathematical context to facilitate our variation generation. More details on the distribution per topic in these two datasets is shown in Table \ref{tab:distributiontopic}.
In total, we have 130 problems as seed data, 65 from the MATH dataset and 65 from the scraped math problems(AoPS). Going forward, we will address the two datasets as: MATH data and AoPS data.

\subsection{Variation Generation}
\label{sec:experimental_design}
Initially, we select problems from the MATH dataset as our seed data. The criteria for selecting this data was such that every instance should have both a mathematical variable and a non-mathematical context. From these, we initially crafted \textbf{four types of variations}, each targeting a distinct linguistic or symbolic component of the problem. 

All variations that we have preserve 1) The numerical values 2) The solution logic 3) The final answer. This is the key differentiation in the variants that we produce from the already existing variant benchmarks. 
\begin{figure}
    \centering
    \includegraphics[width=\linewidth]{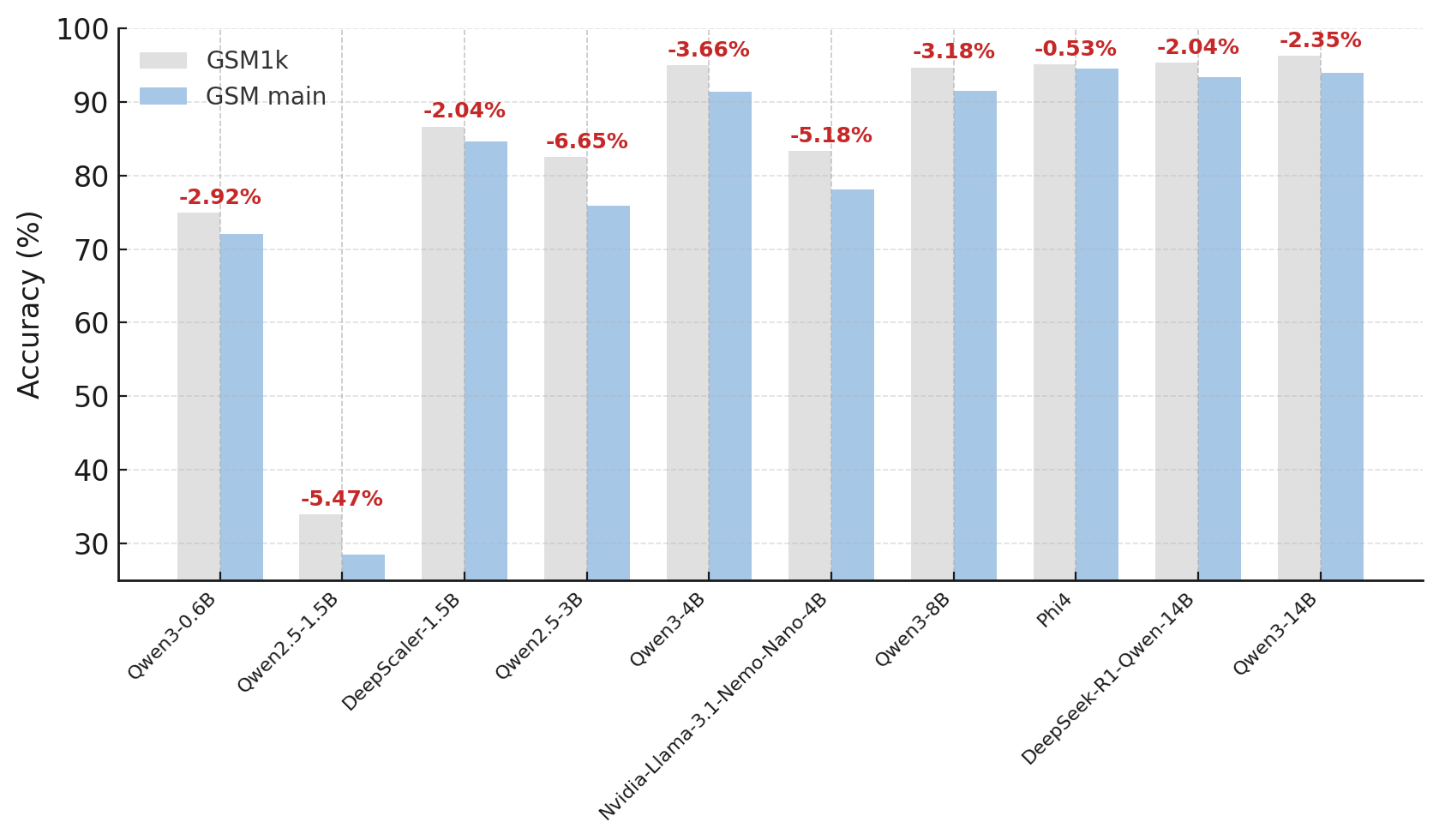}
    \caption{Preliminary experiment to check accuracy drops on the GSM symbolic dataset (GSM Main) as compared to the test set of GSM8k (GSM1k). More details are in Table \ref{tab:gsm_model_performance}.}
    \label{fig:gsmdrops}
\end{figure}
Below is a detailed explanation of every type of variation that we produce:
\par \textbf{Variant 1: Variable Substitution:} 
    Only the variable(s) used in the question are changed. Language, numbers, and context remain unchanged.\\
    \textit{Example:} \texttt{``Harry has $x$ dogs''} $\rightarrow$ \texttt{``Harry has $y$ dogs''}
\par \textbf{Variant 2: Context Substitution:}
    Only the non-mathematical context is changed. Variables and numbers are kept intact.\\
    \textit{Example:} \texttt{``4 dogs''} $\rightarrow$ \texttt{``4 books''}
\par \textbf{Variant 3: Full Rewriting with Variable Fixed:}
    The entire question is reformulated with different language, but the same variable(s) and numerical values are retained.\\
    \textit{Example:} \texttt{``Harry has $x$ dogs''} $\rightarrow$ \texttt{``Oscar sold $x$ books''}
\par \textbf{Variant 4: Full Rewriting with Variable Change:} Both language and variable(s) are changed. Only the numbers and final answer remain fixed.\\
    \textit{Example:} \texttt{``Harry has $x$ dogs''} $\rightarrow$ \texttt{``Oscar sold $y$ books''}

We achieve this using the following two rules: 1) An explicit mathematical unknown (e.g., $x$, $y$, $\theta$). Constants such as $\pi$ are not modified. 2) A non-mathematical real-world element (e.g., ``projectile'', ``dog walking business''). An example of Variant 4 and its corresponding answer is shown in Figure \ref{fig:variation4example}



We ensure that each selected problem contains both a variable and a real-world context to enable controlled and consistent variations across all four types. Before scaling the dataset we do a preliminary experiment to check which variation performs the best. We check model performance across all 4 variations and report an average of 5 runs for one example. Top performing models have the least performance for variation 4. More details are in the Table \ref{tab:model_performance_initial} in the Appendix. Variation 4 is also a combination of all the previous 3 variations. Keeping this in mind, in our future experiments, we focus only on for Variation 4 for comparison going forward.




As explained in section \S\ref{sec:seed_dataset}, we have a total of 130 seed problems. To generate variations per example, we pass the original question, original answer, along with 4-shot examples of original questions and their variations, together with some system instructions to a model. We use Claude-sonnet and Gemini-2.5-flash to generate variations. We generate 4 variation sets, two from Claude and two from Gemini, for each example. Thus, we have a total of 520 variations. This number is comparable to the MATH-500 test set \cite{lightman2023lets}.
We also manually check the generated variations to verify if all the conditions are met and make suitable changes wherever necessary.
\begin{figure*}[ht]
    \centering
    \includegraphics[width=0.7\linewidth]{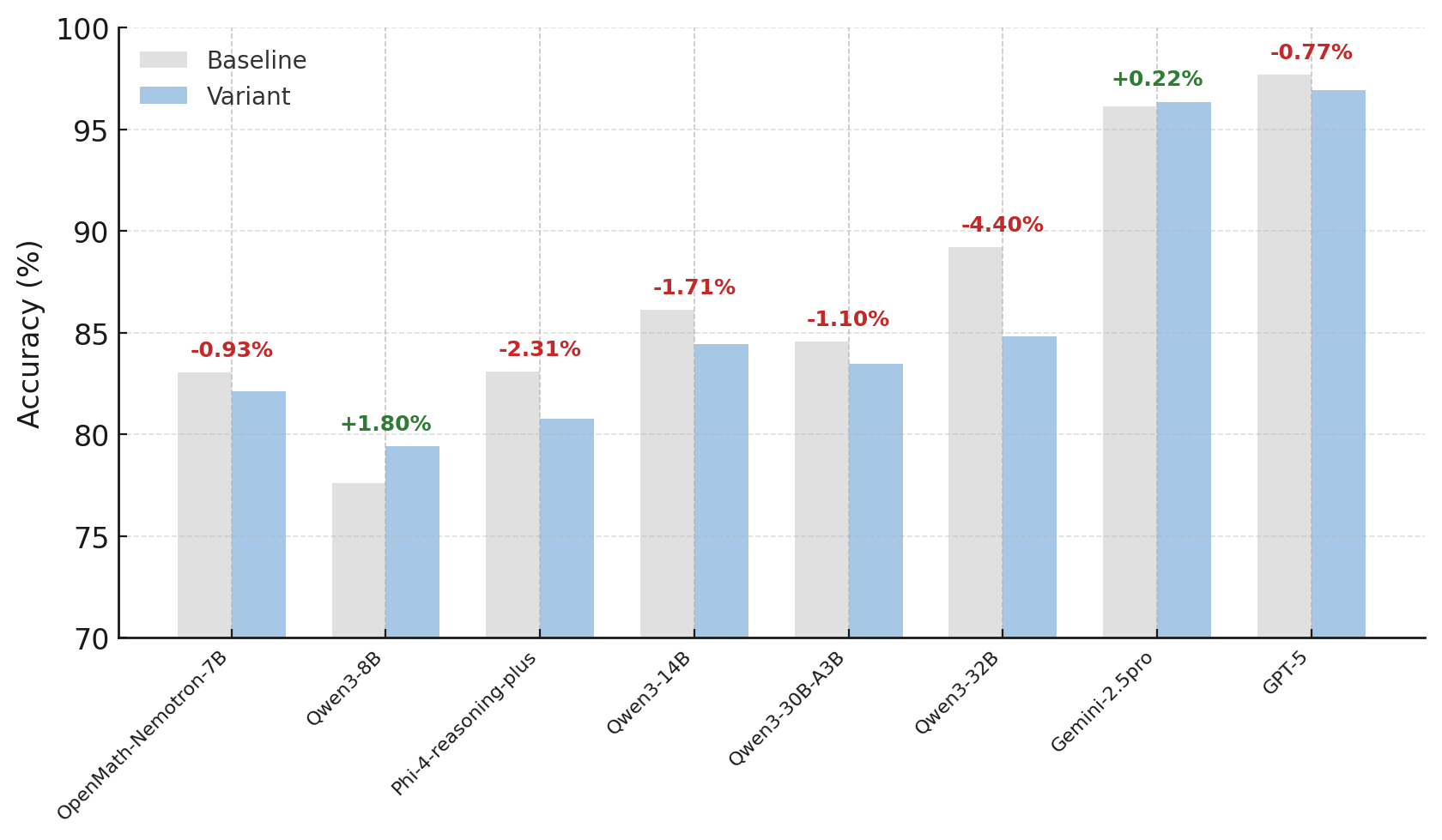}
    \caption{Accuracy drops for some notable models on our variation data.}
    \label{fig:mathdrops}
\end{figure*}


\section{Experimental Setup}
\par\textbf{Models.} We test on a total of 34 model, both open-sourced and closed source, of various sizes and types. We predominantly test on the following families of models: GPT, Llama, Qwen, Phi, Nvidia, Gemini, Claude. \par
\textbf{Evaluation.} All the evaluations are done in a zero-shot setting. We compute the accuracies of the model on the test set on the standard pass@1. Besides that we also calculate the accuracy drop, which is the difference between the variation accuracy and the original accuracy. A negative drop indicates that the model performed poorly on the variants as compared to the original dataset. This indicates a lack of robustness. \par

While evaluating the model, we give it an instruction to output the final answer in a boxed format (e.g. \boxed{final\_answer}). To extract this final answer from the generated output we employ a hybrid strategy of using a function and an LLM. For open source models our hyperparameters are as follows: temperature: 0.6, top\_p : 0.95, max\_output\_len is 14k.
We have a deterministic reward function to check if the generation is in the correct format. If the generation is in a correct format, we use this function to get the final answer and compare it with the ground truth. If the format is incorrect, there still might be cases of the answer being correct but being in the wrong format (e.g. \textit{Answer: final\_answer}). For such edge cases we use an LLM (Gemini-2.5-pro \cite{comanici2025gemini}) as a judge \cite{lambert2024rewardbench} and use that to evaluate the generation and extract the final answer. 


\section{Results}
\subsection{Preliminary Results}


To gauge how models react to variations, we test various latest models on the GSM-symbolic dataset \cite{mirzadeh2024gsm}. Figure \ref{fig:gsmdrops} shows the drop in accuracy of the GSM-symbolic dataset in comparison to GSM8k. Table \ref{tab:gsm_model_performance} has results for the GSMmain, GSM-SYM-P1, GSM-SYM-P2 and the GSM1k dataset. GSM-SYM is the original benchmark dataset where problem wording and numbers are varied, but the final logic remains the same. GSM-SYM-P1 is a dataset which adds 1 extra reasoning step to the answer and GSM-SYM adds two reasoning steps to the answer. GSM1k is the test set of the GSM8k dataset. We report the drop in accuracies between GSM main and GSM1k. All the models exhibit a drop in performance when tested on the benchmark. We do this experiment to get a sense of the drops we should expect when tested on our benchmark.



\subsection{Results on MathRobust-LV data} 
Across all the models, we report accuracy for the baseline, variants and the drop between these two accuracies in Table \ref{tab:final_table}. A negative drop indicates that the model performs better on the baseline than the variant, which indicates a lack of robustness. However, we must note that if a model has a positive increase while the base accuracy is low is not an indicator for the model being robust. Some drop patterns are also shown in Figure \ref{fig:mathdrops}.

\subsubsection{General Trends}
As seen in Table \ref{tab:final_table}, we see that across the models, linguistic variation generally reduces model performance. Most open-source models experience drops of 2–6\% between the baseline and variant settings, suggesting a lack of invariance to surface-level rephrasings. The effect is especially severe for smaller checkpoints such as \texttt{OpenMath-Nemotron-1.5B} (–11.52\%) and \texttt{DeepSeek-R1-Distill-Qwen-1.5B} (–9.59\%). 

We evaluate our dataset on a wide variety of models of different sizes and types. The best model which balances robustness and strong variant performance is \texttt{Qwen3-8B}. Some models exhibit positive increase but perform poorly on variants overall, whereas once variant accuracy surpasses 80\%, the drop is consistently negative.
\begin{figure}[ht]
\centering
\begin{tcolorbox}[onecolqual]
\begingroup\footnotesize\RaggedRight

\textbf{Original Problem.}\\[2pt]
Professor Gamble buys a lottery ticket, which requires that he pick six different integers from $1$ through $46$, inclusive. He chooses his numbers so that the sum of the base-ten logarithms of his six numbers is an integer. It so happens that the integers on the winning ticket have the same property— the sum of the base-ten logarithms is an integer. What is the probability that Professor Gamble holds the winning ticket?\\[4pt]
$\textbf {(A)}\ 1/5\quad
\textbf {(B)}\ 1/4\quad
\textbf {(C)}\ 1/3\quad
\textbf {(D)}\ 1/2\quad
\textbf {(E)}\ 1$\\[6pt]
\textbf{Model (Qwen3-4B} predicts \(\mathbf{B}=1/4\) (correct) \\
\emph{Commentary: Correctly recognizes the 10-smooth pool $\{1,2,4,5,8,10,16,20,25,32,40\}$ and counts 4 valid 6-sets whose product is a power of 10.}\\[8pt]

\hrule height0.35pt \vspace{8pt}

\textbf{Variant Problem.}\\[2pt]
Dr. Chen, a mathematics professor, participates in a special academic lottery where participants must select six distinct positive integers from the range $1$ to $46$, inclusive. Dr. Chen has a particular strategy: she chooses her numbers such that when she calculates $\log_{10}(a_1) + \log_{10}(a_2) + \log_{10}(a_3) + \log_{10}(a_4) + \log_{10}(a_5) + \log_{10}(a_6)$ where $a_1, a_2, a_3, a_4, a_5, a_6$ are her chosen numbers, the result is a whole number.

The lottery officials announce that the winning combination also has this same mathematical property—the sum of the base-ten logarithms of the six winning numbers equals an integer.

Given that both Dr. Chen's ticket and the winning ticket satisfy this logarithmic constraint, what is the probability that Dr. Chen holds the winning ticket?\\[4pt]
$\textbf {(A)}\ 1/5\quad
\textbf {(B)}\ 1/4\quad
\textbf {(C)}\ 1/3\quad
\textbf {(D)}\ 1/2\quad
\textbf {(E)}\ 1$\\[6pt]
\textbf{Model (Qwen3-4B:} predicts \(\mathbf{D}=1/2\) (wrong) \\
\emph{Commentary: Same pool identified, but miscounts valid 6-sets (2 instead of 4), yielding the wrong probability.}\\[8pt]
\par\endgroup
\end{tcolorbox}
\caption{Illustrative example where a model gets the answer of the original question correct but fails on the variation. The two problems are mathematically equivalent (only surface wording differs). The model succeeds on the original but fails on the variant, illustrating brittleness under benign rephrasings.
}
\label{fig:qual-onecol}
\end{figure}

\subsubsection{Scaling Effects}
Parameter scaling improves both absolute accuracy and robustness, but the gains are not linear. Sub-2B models perform poorly (30–60\% baseline) and degrade sharply under variation. Mid-sized models in the 4–15B range achieve more stable baselines (70–85\%) but still drop substantially under stress. Large models in the 30–32B class reach 85–90\% with smaller relative declines (–1\% to –5\%), while scaling further to 70B parameters yields only marginal robustness improvements, suggesting diminishing returns beyond this range. 
\begin{table*}[ht]
\centering
\small
\begin{tabular}{lccccc}
\toprule
\textbf{Model} & \textbf{Params (B)} & \textbf{MATH65} & \textbf{AoPS65} & \textbf{MATH260} & \textbf{AoPS260} \\
\midrule
DeepSeek-R1-Distill-Qwen-1.5B & 1.78 & 86.10\% & 30.77\% & 78.07\% & 19.62\% \\
OpenMath-Nemotron-1.5B        & 1.54 & 86.10\% & 64.62\% & 80.76\% & 46.92\% \\
DeepSeek-R1-Distill-Qwen-7B   & 7.62 & 93.84\% & 52.31\% & 85.77\% & 48.46\% \\
Qwen3-1.7B                    & 2.03 & 86.01\% & 32.31\% & 82.31\% & 31.54\% \\
Qwen3-4B                      & 4.02 & 92.25\% & 72.31\% & 91.54\% & 62.69\% \\
Claude-opus                   & --   & 93.85\% & 56.90\% & 93.44\% & 50.77\% \\
\bottomrule
\end{tabular}
\caption{Comparison of accuracy on MATH vs. AoPS problems for selected models where discrepancies are most pronounced. AoPS subsets are consistently harder, with drops up to 50 points.}
\label{tab:ablation}
\end{table*}
\subsubsection{Architectural Differences}
We also find pronounced differences across model families. Within the Qwen family, larger models such as \texttt{Qwen3-32B} achieve the highest open-source baselines (89.21\%) but still lose robustness (–4.40\%). The “Thinking” checkpoints (e.g., \texttt{Qwen3-30B-Thinking}) generally drop more than their non-thinking counterparts. 
Distilled DeepSeek models scale competitively, yet their smallest versions remain fragile. NVIDIA’s Nemotron family shows similar behavior: the 7B and 8B checkpoints are strong, but the 1.5B variant collapses under variation (–11.52\%). Microsoft’s \texttt{Phi-4} reasoning models (14.7B) strike a balance, offering high baselines and moderate drops, whereas LLaMA-3.1 Instruct trails substantially (21.39\% baseline), underscoring the limits of generic instruction tuning in mathematical reasoning. 

\subsubsection{Closed source models}
We evaluate five widely used closed-source models: Gemini-2.5-pro, Gemini-2.5-flash, Claude-sonnet, Claude-opus, GPT-5. Gemini-2.5-pro is the best performing model with a positive increase, followed by GPT-5 which has a drop less than one percent. Both the Claude models show a drop of 3.27 percent.

\section{Additional Experiments and Observations}
\paragraph{MATH vs. AoPS dataset discrepancies.} 
The comparison between \emph{MATH test} and \emph{AoPS scraped} subsets reveals large and systematic gaps as shown in Table \ref{tab:ablation}.
For several models, accuracy on MATH problems exceeds 85–90\%, while performance on AoPS problems drops by 30–50 points. 
For example, \texttt{DeepSeek-R1-Distill-Qwen-1.5B} falls from 86.10\% to 30.77\% on the 65-problem subsets, and from 78.07\% to 19.62\% on the 260-problem variants. 
Similarly, \texttt{Qwen3-1.7B} and \texttt{DeepSeek-R1-Distill-Qwen-7B} each lose more than 40 points moving from MATH to AoPS. 
Even stronger models such as \texttt{Qwen3-4B} and closed-source \texttt{Claude-Opus} exhibit significant degradation, despite robust performance on MATH test. 
These results suggest that AoPS problems introduce linguistic and structural diversity that stresses generalization in ways the MATH test does not, exposing robustness gaps even in high-capacity models. The low accuracy on the AoPS dataset for Claude models is also consistent with the performance of the models on the AIME 2025 dataset which has similar data like the AoPS.\footnote{\url{https://www.vals.ai/benchmarks/aime-2025-09-08}}.

\paragraph{Qualitative example.}

To further illustrate the brittleness we quantify in aggregate metrics, we present paired qualitative examples where surface-level changes cause inconsistent outcomes. In Figure~\ref{fig:qual-onecol}, both the original and variant questions encode the same mathematical structure, requiring the solver to recognize that the product of the chosen numbers must be a power of ten. The only differences are cosmetic, entity names and phrasing such as “whole number” vs.\ “integer.” Yet, the model succeeds on the original but fails on the variant, miscounting the valid subsets and producing the wrong probability. This contrast highlights that the model is not failing at mathematics per se, but rather at maintaining stable reasoning across semantically equivalent formulations. Such case studies demonstrate how linguistic sensitivity, rather than mathematical hardness, often drives errors, reinforcing the need for variation-based evaluation.

\section{Conclusion}
Benchmark performance alone is insufficient to establish evidence of genuine mathematical reasoning. By systematically decoupling numerical structure from linguistic surface form, MathRobust-LV uncovers latent brittleness across model families and parameter scales. We anticipate that these resources will facilitate subsequent research on contamination-resistant benchmarks, tool-augmented reasoning methodologies, and architectures capable of faithfully manipulating formal abstractions.

\section{Future Work and Limitations}

While our evaluation methodology is appropriate for practical educational use cases,
our study does have its limitations. First, the benchmark is restricted to 130 seed problems due to strict selection criteria requiring both a variable and contextual information. While this ensures controlled variation, it limits the diversity of problem types. Second, although we preserve numerical structure and answers, models may still underperform due to exposure to unfamiliar surface forms or contexts, making it difficult to isolate linguistic robustness from generalization ability. Third, our evaluations rely solely on pass@1, which does not capture nuanced reasoning chains or partially correct outputs that could offer a more graded view of robustness.

For future work, we aim to extend MathRobust-LV by incorporating a wider range of problem types and domains such as physics and finance. To scale up, we will explore automatic rewriters for large-scale variant generation and conduct fine-grained error analyses to identify specific linguistic or symbolic triggers for failure. Additionally, we intend to examine whether fine-tuning models with a RL setting on variant-rich datasets enhances robustness.



\section{Ethics Statement}
This work evaluates the robustness of large language models (LLMs) to linguistic perturbations in math reasoning tasks. All models tested are publicly available through open-source repositories or APIs, and no proprietary or private data were used during evaluation. 


While our benchmark highlights brittleness in LLMs, we do not claim definitive insights into model internals or cognitive capabilities. Our reworded variations are designed for research purposes and should not be construed as adversarial attacks or tools for model manipulation.


This research does not involve human subjects or personal data. We encourage future work to explore similar robustness evaluations in languages beyond English and to assess potential fairness or equity implications of linguistic brittleness in AI systems.

\bibliography{custom}

\appendix
\section{Preminilary results across variations}
\begin{table*}
\centering
\begin{tabular}{|l|c|c|c|c|c|}
\hline
\textbf{Model} & \textbf{Original} & \textbf{Variation 1} & \textbf{Variation 2} & \textbf{Variation 3} & \textbf{Variation 4} \\
\hline
o1-mini & 100\% & 100\% & 100\% & 0\% & 0\% \\
o1-preview & 100\% & 100\% & 100\% & 100\% & 60\% \\
GPT-4o-mini & 100\% & 100\% & 100\% & 80\% & 20\% \\
GPT-4o & 100\% & 100\% & 100\% & 0\% & 20\% \\
GPT-4 & 0\% & 0\% & 0\% & 0\% & 0\% \\
Llama3-8B & 0\% & 0\% & 0\% & 0\% & 0\% \\
Llama3-70B & 40\% & 0\% & 0\% & 0\% & 0\% \\
Llama 3.1-8B & 100\% & 0\% & 0\% & 0\% & 0\% \\
Llama 3.1-70B & 40\% & 60\% & 100\% & 100\% & 0\% \\
Llama 3.1-405B & 100\% & 80\% & 0\% & 0\% & 0\% \\
Deepseek-coder-v2 & 60\% & 80\% & 40\% & 20\% & 0\% \\
Deepseek-chat-v2 & 40\% & 40\% & 0\% & 60\% & 0\% \\
Deepseek-v2.5 & 60\% & 80\% & 40\% & 0\% & 0\% \\
Phi-3.5-mini & 40\% & 0\% & 40\% & 60\% & 0\% \\
Phi3-medium-128k & 0\% & 0\% & 20\% & 0\% & 0\% \\
Mistral-Nemo-12B & 20\% & 0\% & 0\% & 0\% & 0\% \\
\hline
\end{tabular}
\caption{Preliminary results across different model variations to understand which variation to choose from. Variation 4 is the one chosen. Results are the average of 5 runs for a single example.}
\label{tab:model_performance_initial}
\end{table*}

\begin{table*}[t]
\centering
\small
\begin{tabular}{|p{0.12\linewidth}|p{0.83\linewidth}|}
\hline
\textbf{Version} & \textbf{Full Question Text} \\
\hline
\textbf{Original} &
Denali and Nate work for a dog walking business and are paid for each dog they walk. Denali is responsible for 16 dogs and Nate is responsible for 12 dogs. Under the company’s new policy, they will be assigned or unassigned new dogs in groups of \( x \) dogs. The ratio of Denali’s pay to Nate’s pay would be the same if Denali started walking \( x \) more dogs and Nate stayed at 12 dogs or if \( x \) of Nate’s dogs were reassigned to Denali. Find \( x \) if \( x \neq 0 \). \\
\hline
\textbf{Variation 1} &
Denali and Nate work for a dog walking business and are paid for each dog they walk. Denali is responsible for 16 dogs and Nate is responsible for 12 dogs. Under the company’s new policy, they will be assigned or unassigned new dogs in groups of \( y \) dogs. The ratio of Denali’s pay to Nate’s pay would be the same if Denali started walking \( y \) more dogs and Nate stayed at 12 dogs or if \( y \) of Nate’s dogs were reassigned to Denali. Find \( y \) if \( y \neq 0 \). \\
\hline
\textbf{Variation 2} &
Denali and Nate work in a library and are responsible for cataloging a certain number of books. Denali is responsible for 16 books and Nate is responsible for 12 books. Under the library’s new policy, they will be assigned or unassigned new books in groups of \( x \) books. The ratio of Denali’s work to Nate’s work would be the same if Denali started cataloging \( x \) more books and Nate stayed at 12 books or if \( x \) of Nate’s books were reassigned to Denali. Find \( x \) if \( x \neq 0 \). \\
\hline
\textbf{Variation 3} &
Farmer Alice and Farmer Bob each grow a different type of crop. Alice currently has 16 units of her crop, and Bob has 12 units of his. Under certain agricultural changes, the ratio of Alice’s crop to Bob’s crop would remain the same if Alice were to harvest \( x \) more units of her crop and Bob’s crop yield stayed the same, or if Bob were to give away \( x \) units of his crop to Alice. Find the value of \( x \) if \( x \neq 0 \). \\
\hline
\textbf{Variation 4} &
Alice and Bob are avid book collectors. Alice currently has 16 rare first editions, while Bob has 12 signed copies. They have agreed to trade books in groups of y books where y is a positive integer. The ratio of Alice’s first editions to Bob’s signed copies would remain the same if Alice acquired 4y more first editions and Bob kept his collection unchanged, or if Bob gave away y of his signed copies to Alice. Find the value of y if y is not 0. \\
\hline
\end{tabular}
\caption{Full text of the original question and its three surface variations. While the domains vary (dog walking, book cataloging, crop yield), the underlying mathematical relationship and variable structure remain aligned.}
\label{tab:variation-full-text}
\end{table*}

\begin{table*}
\centering
\small
\begin{tabular}{|l|c|c|c|c|c|c|}
\hline
\textbf{Model} & \textbf{Params (B)} & \textbf{GSM1k} & \textbf{GSM main} & \textbf{Drop} & \textbf{GSM p1} & \textbf{GSM p2} \\
\hline
Qwen3-0.6B & 0.75 & 74.98\% & 72.06\% & 2.92\% & 58.24\% & 33.39\% \\
Nvidia-OpenMath-Nemo-1.5B & 1.54 & 78.76\% & 78.53\% & 0.23\% & 66.05\% & 48.39\% \\
Qwen2.5-1.5B & 1.54 & 33.99\% & 28.52\% & 5.47\% & 26.06\% & 13.37\% \\
Qwen2.5-Math-1.5B & 1.54 & 84.29\% & 79.36\% & 4.93\% & 67.28\% & 50.48\% \\
DeepScaler-1.5B & 1.78 & 86.65\% & 84.61\% & 2.04\% & 76.99\% & 60.90\% \\
Qwen-3-1.7B & 2.03 & 89.84\% & 87.50\% & 2.34\% & 82.05\% & 71.11\% \\
Qwen2.5-3B & 3.09 & 82.56\% & 75.91\% & 6.65\% & 67.60\% & 44.08\% \\
Phi4-mini & 3.84 & 89.16\% & 86.46\% & 2.70\% & 79.17\% & 63.54\% \\
Phi4-mini-reasoning & 3.84 & 88.55\% & 91.52\% & -2.97\% & 90.09\% & 85.31\% \\
Qwen3-4B & 4.02 & 95.00\% & 91.34\% & 3.66\% & 88.46\% & 83.10\% \\
Nvidia-Llama-3.1-Nemo-Nano-4B & 4.51 & 83.32\% & 78.14\% & 5.18\% & 68.53\% & 50.76\% \\
Nvidia-AceMath-RL-Nemo-7B & 7.62 & 92.64\% & 89.00\% & 3.64\% & 86.59\% & 84.23\% \\
Nvidia-OpenMath-Nemo-7B & 7.62 & 88.86\% & 87.21\% & 1.65\% & 81.78\% & 76.52\% \\
Nvidia-Llama-3.1-Nemo-Nano-8B & 8.03 & 79.98\% & 80.02\% & -0.04\% & 80.04\% & 78.71\% \\
{DeepSeek-R1-0528-Qwen-8B} & 8.19 & 93.40\% & 91.65\% & 1.75\% & 86.86\% & 83.44\% \\
Qwen3-8B & 8.19 & 94.69\% & 91.51\% & 3.18\% & 90.58\% & 87.90\% \\
Phi4 & 14.7 & 95.07\% & 94.54\% & 0.53\% & 91.60\% & 88.32\% \\
Phi4-reasoning & 14.7 & 90.70\% & 90.31\% & 0.39\% & 91.20\% & 91.10\% \\
Phi4-reasoning-plus & 14.7 & 83.36\% & 82.57\% & 0.79\% & 82.87\% & 84.68\% \\
DeepSeek-R1-Qwen-14B & 14.8 & 95.38\% & 93.34\% & 2.04\% & 92.16\% & 90.56\% \\
Qwen3-14B & 14.8 & 96.29\% & 93.94\% & 2.35\% & 92.94\% & 89.36\% \\
\hline
\end{tabular}
\caption{Performance of various models on GSM1k and GSM-Symbolic variants (main, p1, p2). Drop = GSM1k – GSM main.}
\label{tab:gsm_model_performance}
\end{table*}

\begin{table*}
\centering
\resizebox{\textwidth}{!}{
\begin{tabular}{lcccccccc}
\toprule
\multicolumn{9}{c}{\textbf{Open-Source Models}} \\
\midrule
\textbf{Model} & \textbf{Params (B)} & \textbf{MATH65} & \textbf{AoPS65} & \textbf{Baseline130} & \textbf{MATH260} & \textbf{AoPS260} & \textbf{Variant520} & \textbf{Drop} \\
\midrule
agentica-org\_DeepScaleR-1\_5B-Preview & 1.78 & 84.42\% & 33.85\% & 59.13\% & 79.62\% & 31.92\% & 55.77\% & -3.36\% \\
deepseek-ai\_DeepSeek-R1-0528-Qwen3-8B & 8.19 & 76.77\% & 60.00\% & 68.39\% & 78.07\% & 56.92\% & 67.50\% & -0.89\% \\
deepseek-ai\_DeepSeek-R1-Distill-Llama-70B & 70.6 & 92.25\% & 73.85\% & 83.05\% & 92.69\% & 70.38\% & 81.54\% & -1.52\% \\
deepseek-ai\_DeepSeek-R1-Distill-Llama-8B & 8.03 & 86.01\% & 44.62\% & 65.31\% & 82.69\% & 47.31\% & 65.00\% & -0.31\% \\
deepseek-ai\_DeepSeek-R1-Distill-Qwen-1\_5B & 1.78 & 86.10\% & 30.77\% & 58.44\% & 78.07\% & 19.62\% & 48.85\% & -9.59\% \\
deepseek-ai\_DeepSeek-R1-Distill-Qwen-32B & 32.8 & 95.34\% & 64.62\% & 79.98\% & 93.46\% & 64.23\% & 78.85\% & -1.13\% \\
deepseek-ai\_DeepSeek-R1-Distill-Qwen-7B & 7.62 & 93.84\% & 52.31\% & 73.08\% & 85.77\% & 48.46\% & 67.12\% & -5.96\% \\
meta-llama\_Llama-3\_1-8B-Instruct & 8.03 & 32.08\% & 10.70\% & 21.39\% & 44.22\% & 4.23\% & 24.23\% & +2.83\% \\
microsoft\_phi-4 & 14.7 & 79.76\% & 16.90\% & 48.33\% & 76.15\% & 21.11\% & 48.63\% & +0.30\% \\
microsoft\_Phi-4-mini-instruct & 3.84 & 45.89\% & 7.60\% & 26.75\% & 52.69\% & 4.60\% & 28.65\% & +1.90\% \\
microsoft\_Phi-4-mini-reasoning & 3.84 & 89.18\% & 50.77\% & 69.97\% & 87.31\% & 40.38\% & 63.85\% & -6.13\% \\
microsoft\_phi-4-reasoning & 14.7 & 93.84\% & 69.23\% & 81.54\% & 90.00\% & 66.92\% & 78.46\% & -3.08\% \\
microsoft\_phi-4-reasoning-plus & 14.7 & 93.84\% & 72.31\% & 83.08\% & 91.92\% & 69.62\% & 80.77\% & -2.31\% \\
nvidia\_AceMath-RL-Nemotron-7B & 7.62 & 95.34\% & 63.07\% & 79.20\% & 91.92\% & 62.30\% & 77.11\% & -2.09\% \\
nvidia\_Llama-3\_1-Nemotron-Nano-4B-v1\_1 & 4.51 & 76.68\% & 64.62\% & 70.65\% & 76.92\% & 56.54\% & 66.73\% & -3.92\% \\
nvidia\_Llama-3\_1-Nemotron-Nano-8B-v1 & 8.03 & 90.67\% & 75.38\% & 83.03\% & 90.00\% & 63.85\% & 76.93\% & -6.10\% \\
nvidia\_OpenMath-Nemotron-1\_5B & 1.54 & 86.10\% & 64.62\% & 75.36\% & 80.76\% & 46.92\% & 63.84\% & -11.52\% \\
nvidia\_OpenMath-Nemotron-7B & 7.62 & 92.25\% & 73.85\% & 83.05\% & 91.15\% & 73.08\% & 82.12\% & -0.94\% \\
Qwen\_Qwen3-0\_6B & 0.75 & 61.38\% & 7.69\% & 34.53\% & 66.15\% & 9.23\% & 37.69\% & +3.16\% \\
Qwen\_Qwen3-1\_7B & 2.03 & 86.01\% & 32.31\% & 59.16\% & 82.31\% & 31.54\% & 56.93\% & -2.23\% \\
Qwen\_Qwen3-14B & 14.8 & 95.34\% & 76.92\% & 86.13\% & 93.46\% & 75.38\% & 84.42\% & -1.71\% \\
Qwen\_Qwen3-30B-A3B & 30.5 & 93.75\% & 75.38\% & 84.56\% & 94.23\% & 72.69\% & 83.46\% & -1.10\% \\
Qwen\_Qwen3-30B-A3B-Thinking-2507 & 30.5 & 96.83\% & 75.38\% & 86.10\% & 93.07\% & 68.85\% & 80.96\% & -5.14\% \\
Qwen\_Qwen3-32B & 32.8 & 95.34\% & 83.08\% & 89.21\% & 92.69\% & 76.92\% & 84.81\% & -4.40\% \\
Qwen\_Qwen3-4B & 4.02 & 92.25\% & 72.31\% & 82.28\% & 91.54\% & 62.69\% & 77.12\% & -5.17\% \\
Qwen\_Qwen3-4B-Thinking-2507 & 4.02 & 88.99\% & 61.54\% & 75.26\% & 88.07\% & 56.92\% & 72.50\% & -2.77\% \\
Qwen\_Qwen3-8B & 8.20 & 92.16\% & 63.08\% & 77.62\% & 91.92\% & 66.92\% & 79.42\% & +1.80\% \\
Qwen\_QwQ-32B & 32.5 & 96.92\% & 72.31\% & 84.62\% & 92.69\% & 72.31\% & 82.50\% & -2.12\% \\
Qwen\_QwQ-32B-Preview & 32.5 & 86.01\% & 46.15\% & 66.08\% & 81.92\% & 48.85\% & 65.39\% & -0.69\% \\
POLARIS-Project\_Polaris-4B-Preview & 4.02 & 89.23\% & 58.46\% & 73.85\% & 88.85\% & 50.77\% & 69.81\% & -4.04\% \\
POLARIS-Project\_Polaris-1.7B-Preview & 1.72 & 80.00\% & 40.00\% & 60.00\% & 79.62\% & 32.31\% & 55.97\% & -4.04\% \\
custom-checkpoints\_phase4\_step250 & 4.02 & 87.69\% & 52.31\% & 70.00\% & 89.23\% & 48.46\% & 68.85\% & -1.15\% \\
\midrule
\multicolumn{9}{c}{\textbf{Closed-Source Models}} \\
\midrule
Claude-sonnet & -- & 92.31\% & 55.38\% & 73.85\% & 91.92\% & 49.23\% & 70.58\% & -3.27\% \\
Claude-opus & -- & 93.85\% & 56.90\% & 75.38\% & 93.44\% & 50.77\% & 72.11\% & -3.27\% \\
Gemini-2.5pro & -- & 98.46\% & 93.80\% & 96.13\% & 96.92\% & 95.77\% & 96.35\% & +0.21\% \\
Gemini-2.5flash & -- & 95.38\% & 93.85\% & 94.62\% & 93.85\% & 88.46\% & 91.16\% & -3.46\% \\
GPT-5 & -- & 98.46\% & 96.92\% & 97.69\% & 96.92\% & 96.92\% & 96.92\% & -0.77\% \\
\bottomrule
\end{tabular}
}
\caption{Full results across all models, separated into open-source and closed-source groups. Columns report performance on MATH test vs. AoPS scraped subsets (65 and 260), the combined baseline (130), aggregate variants (520), and drop.}
\end{table*}

\end{document}